\documentclass[sigconf,nonacm]{acmart}
\setcopyright{none}              % Suppresses copyright statement and text

\acmConference[]{}{}{}
\acmBooktitle{}
\acmDOI{}
\acmISBN{}
\acmYear{}
\copyrightyear{}
\usepackage{tikz}
\usetikzlibrary{arrows.meta,calc,positioning,fit,backgrounds}
\usepackage{multirow}
\usepackage{makecell}
\usepackage{booktabs}
\usepackage{graphicx}
\usepackage{siunitx}
\usepackage{tabularx}
\usepackage{placeins}
\usepackage{subcaption}
\usepackage{xurl}
\newcommand{\modelname}{TailDART}
\newcommand{\metricnote}{All metrics are lower-is-better.}


\usepackage[normalem]{ulem}
\usepackage{xcolor}

\graphicspath{{figures/}}

\title{TailDART: Compute-Efficient Foundation Models for Financial Time-Series Forecasting}
\author{Chiang Qin Zhi}
\affiliation{%
  \institution{Nanyang Technological University}
  \country{Singapore}
}

\author{Ranjan Satapathy}
\affiliation{
  \institution{Agency for Science, Technology and Research}
  \country{Singapore}
}

\author{Erik Cambria}
\affiliation{%
  \institution{Nanyang Technological University}
  \country{Singapore}
}

\author{Joyjit Chattoraj}
\affiliation{%
  \institution{Agency for Science, Technology and Research}
  \country{Singapore}
}

\begin{document}
\raggedbottom

\begin{abstract}
Although time-series foundation models (TSFMs) achieve strong zero-shot transferability across high-information-density physical systems, applying overparameterized TSFMs to quantitative finance is fundamentally constrained by low signal-to-noise ratios, non-stationarity, and prohibitive autoregressive decoding latency. To reconcile representation learning with institutional operational constraints, this study presents TailDART, a dense 2.68 million-parameter foundation model engineered for financial time-series forecasting. By integrating grouped-query attention into a lightweight backbone, the architecture drastically reduces Key-Value cache memory bandwidth overhead, achieving major inference acceleration (up to 48.1\%) over prevailing state-of-the-art TSFMs to enable low-latency, high-throughput streaming execution. Furthermore, multi-asset benchmarks across varying temporal resolutions show that a compact model can remain competitive for market forecasting: TailDART matches the predictive fidelity of substantially larger baselines while demonstrating superior error resilience in high-volatility regimes.
\end{abstract}


\maketitle

\section{Introduction}
\label{sec:introduction}
% Starting with general Time-Series Foundation Models (TSFMs) establishes the broader AI context, while quickly narrowing down to finance highlights the domain-specific friction.
Time-series forecasting has undergone a paradigm shift, transitioning from domain-specific statistical estimators~\cite{box1970time, corsi2009simple} and task-specific deep learning models~\cite{gers2000learning, oreshkin2019nbeats, DAI2026114235} to zero-shot Time-Series Foundation Models (\textbf{TSFMs})~\cite{ye2024timeseriesfoundation}. Inspired by the empirical scaling laws and autoregressive pre-training paradigms of Large Language Models (LLMs), modern TSFMs leverage deep Transformer backbones trained on vast, multi-domain temporal corpora~\cite{ansari2024chronos, woo2024moirai}. These architectures achieve remarkable cross-domain transferability, unifying univariate and multivariate prediction tasks under standardized tokenization and pre-training objectives. By capturing generalizable temporal primitives, foundation models eliminate the need for dataset-specific retraining, providing robust zero-shot baselines across physical, industrial, and environmental domains such as energy grids, weather forecasting, and traffic monitoring.

%The Dual-Bottleneck Structure (P2 & P3): Splitting the problem into Predictive Limitations (P2: low SNR, noise overfitting, scaling breakdown) and Operational Limitations (P3: inference latency, memory bandwidth, real-time unviability) ensures your paper addresses both methodological theory and practical systems engineering.
However, directly transferring the overparameterization paradigm of generic TSFMs to financial time-series forecasting exposes severe domain-specific friction. Unlike language or physical sensor tokens, which exhibit high information density and structural syntax, financial return series are governed by extremely low Signal-to-Noise Ratios (SNRs), market non-stationarity, and heavy-tailed, non-Gaussian return distributions~\cite{fama1970efficient, gu2020empirical, sun2026finstressts}. Recent empirical evidence suggests that forecasting performance depends jointly on model capacity and data complexity, challenging one-size-fits-all parameter scaling~\cite{lopez2018advances, lo2025bigger}. This interaction raises a fundamental, unanswered question in quantitative finance: \textit{Does the ``bigger is better''} scaling hypothesis hold for noisy financial time series, or is there a practical model size under a fixed training budget?

% Synthesis via Research Questions (P4): Framing the problem around two distinct bottlenecks allows you to pose two corresponding, tightly focused Research Questions:
        %%RQ1 targets the theoretical capacity/scaling bottleneck (Data & Parameter Scaling).
        %%RQ2 targets the architectural/efficiency bottleneck (Latency & Volatility Performance).
Compounding this predictive challenge is a severe operational inference bottleneck during live deployment. Autoregressive TSFMs operating at scale ($10\,\text{M}$ to $100\,\text{M}+$ parameters) incur prohibitive memory bandwidth overhead during decoding, driven by sequential key-Value cache allocation and multi-head attention projections~\cite{ainslie2023gqa}. In institutional financial infrastructure, such as streaming risk analytics, algorithmic execution, and real-time multi-asset portfolio rebalancing, inference latency budgets are strictly bounded to sub-millisecond windows. While recent TSFMs such as timesfm-2.5 use checkpoints with hundreds of millions of parameters~\cite{google2026timesfm25}, deploying such models through dedicated GPU cloud infrastructure can introduce non-trivial network latency and high operational costs, making inference efficiency consequential for high-throughput forecasting systems~\cite{googlecloud2026gpupricing}.

To bridge the gap between foundation model generalization and the operational constraints of financial markets, this paper addresses two central Research Questions:
\begin{itemize} 
   \item Capacity \& Scaling: Given the bounded entropy and low SNR of regime-consistent financial series, does parameter scaling monotonically improve out-of-sample forecasting, or is there an empirically preferred model size ($N^*$) under the tested training setup?
    \item Latency \& Volatility Efficiency: Can a dense, compute-optimal architecture eliminate decoding latency bottlenecks via memory-efficient attention mechanisms while maintaining superior accuracy across diverse asset classes and extreme volatility regimes?
\end{itemize}

To reconcile the trade-off between foundation model representation learning and the predictive and operational constraints of institutional quantitative finance, this study presents \textbf{TailDART}, a compute-efficient foundation model engineered for low-latency, multi-asset returns forecasting. The primary contributions of this work are summarized as follows:

\begin{enumerate}
    \item Model Capacity Efficiency: This paper demonstrates that a dense architecture ($\approx$~\textbf{2.68 M} parameters) can remain highly competitive with significantly larger TSFMs, showing that massive parameter scaling is not strictly necessary for effective financial forecasting.
    \item Inference Latency Reduction: The proposed architecture incorporates grouped-query attention (\textbf{GQA}) within a lightweight backbone, substantially lowering key-Value cache memory bandwidth demands and markedly reducing inference latency (48.1\%) compared to state-of-the-art TSFMs.
    \item Empirical Multi-Asset Evaluation: An empirical evaluation across diverse asset classes and temporal frequencies indicates that \modelname{} delivers solid predictive performance, displaying notable robustness during high-volatility market conditions.
\end{enumerate}

\FloatBarrier
\begin{figure}[t]
  \centering
  \scalebox{0.85}{%
  \begin{tikzpicture}[
    node distance=2.4mm and 3mm,
    font=\scriptsize,
    flow/.style={-{Latex[length=1.5mm]}, semithick},
    skip/.style={-{Latex[length=1.5mm]}, semithick, black!45},
    io/.style={draw=ACMBlue!65, rounded corners=2pt, align=center, fill=ACMBlue!8,
      minimum height=6mm, text width=0.74\columnwidth, inner ysep=1.3mm},
    op/.style={draw=black!45, rounded corners=2pt, align=center, fill=black!7,
      minimum height=5.5mm, text width=0.60\columnwidth, inner ysep=1.2mm},
    norm/.style={draw=ACMOrange!75, rounded corners=3pt, align=center, fill=ACMOrange!16,
      minimum height=3.4mm, text width=0.40\columnwidth, font=\tiny, inner ysep=0.5mm},
    choice/.style={draw=ACMPurple!60, rounded corners=2pt, align=center, fill=ACMPurple!12,
      minimum height=6mm, text width=0.28\columnwidth, inner ysep=1.1mm},
    add/.style={draw=black!55, circle, inner sep=0pt, minimum size=3.4mm, fill=white},
    lbl/.style={font=\tiny\itshape, text=ACMDarkBlue, align=left}
  ]
    % --- input pipeline ---
    \node[io] (input) {Series context $y_{t-W+1:t}$ and time features $\mathbf{x}$};
    \node[io, below=of input] (features) {Robust window scaling $\rightarrow$ lag extraction\\
      concatenate location, scale, and six time features};
    \node[io, below=of features] (projection) {Linear input projection};

    % --- decoder block: attention sublayer ---
    \node[norm, below=7mm of projection] (an1) {RMSNorm};
    \node[op, below=of an1] (attn) {Grouped-query causal attention + RoPE};
    \node[norm, below=of attn] (an2) {RMSNorm};
    \node[add, below=2.4mm of an2] (add1) {$+$};

    % --- decoder block: feed-forward sublayer ---
    \node[norm, below=2.8mm of add1] (mn1) {RMSNorm};
    \node[choice, below=3mm of mn1, xshift=-0.185\columnwidth] (dense) {\textbf{Dense}\\GeLU-gated MLP};
    \node[choice, below=3mm of mn1, xshift=0.185\columnwidth] (moe) {\textbf{MoE}\\top-1 router, 3 experts};
    \coordinate (mmerge) at ($(dense.south)!0.5!(moe.south)$);
    \node[norm, below=3mm of mmerge] (mn2) {RMSNorm};
    \node[add, below=2.4mm of mn2] (add2) {$+$};

    % --- output head ---
    \node[io, below=4mm of add2] (head) {Final RMSNorm $\rightarrow$ distribution parameters\\
      Student's $t$ (primary) or Johnson's $\text{S}_U$};
    \node[io, below=of head] (output) {Restore window location and scale\\$p_\theta(y_{t+1}\mid y_{t-W+1:t},\mathbf{x})$};

    % --- main flow ---
    \draw[flow] (input) -- (features);
    \draw[flow] (features) -- (projection);
    \draw[flow] (projection) -- (an1);
    \draw[flow] (an1) -- (attn);
    \draw[flow] (attn) -- (an2);
    \draw[flow] (an2) -- (add1);
    \draw[flow] (add1) -- (mn1);
    \draw[flow] (mn1.south) -- ++(0,-1.2mm) -| (dense.north);
    \draw[flow] (mn1.south) -- ++(0,-1.2mm) -| (moe.north);
    \draw[flow] (dense.south) -- ++(0,-1.2mm) -| (mn2.north);
    \draw[flow] (moe.south) -- ++(0,-1.2mm) -| (mn2.north);
    \draw[flow] (mn2) -- (add2);
    \draw[flow] (add2) -- (head);
    \draw[flow] (head) -- (output);

    % --- residual skip connections ---
    \draw[skip] (an1.north) -- ($(an1.north)+(-0.40\columnwidth,0)$) |- (add1.west);
    \draw[skip] (mn1.north) -- ($(mn1.north)+(-0.40\columnwidth,0)$) |- (add2.west);

    % --- decoder-block container ---
    \coordinate (bl) at ($(an1.north)+(-0.42\columnwidth,3mm)$);
    \coordinate (br) at ($(moe.east)+(0.03\columnwidth,0)$);
    \begin{scope}[on background layer]
      \node[draw=ACMBlue!45, dashed, rounded corners=4pt, fill=ACMBlue!4,
        fit=(bl)(br)(an1)(add2)(dense)(moe), inner sep=2.4mm] (block) {};
    \end{scope}
    \node[lbl, anchor=north west, xshift=1.2mm, yshift=-1.6mm]
      at (block.north west) {$\times\,N$ decoder blocks};
  \end{tikzpicture}%
  }
  \caption{\modelname{} architecture. Each decoder block applies pre- and post-RMSNorm with a residual add around a grouped-query attention sublayer and a feed-forward sublayer. The dense model uses a GeLU-gated MLP; the MoE variant routes each token to one of three experts (top-1).}
  \label{fig:architecture}
\end{figure}
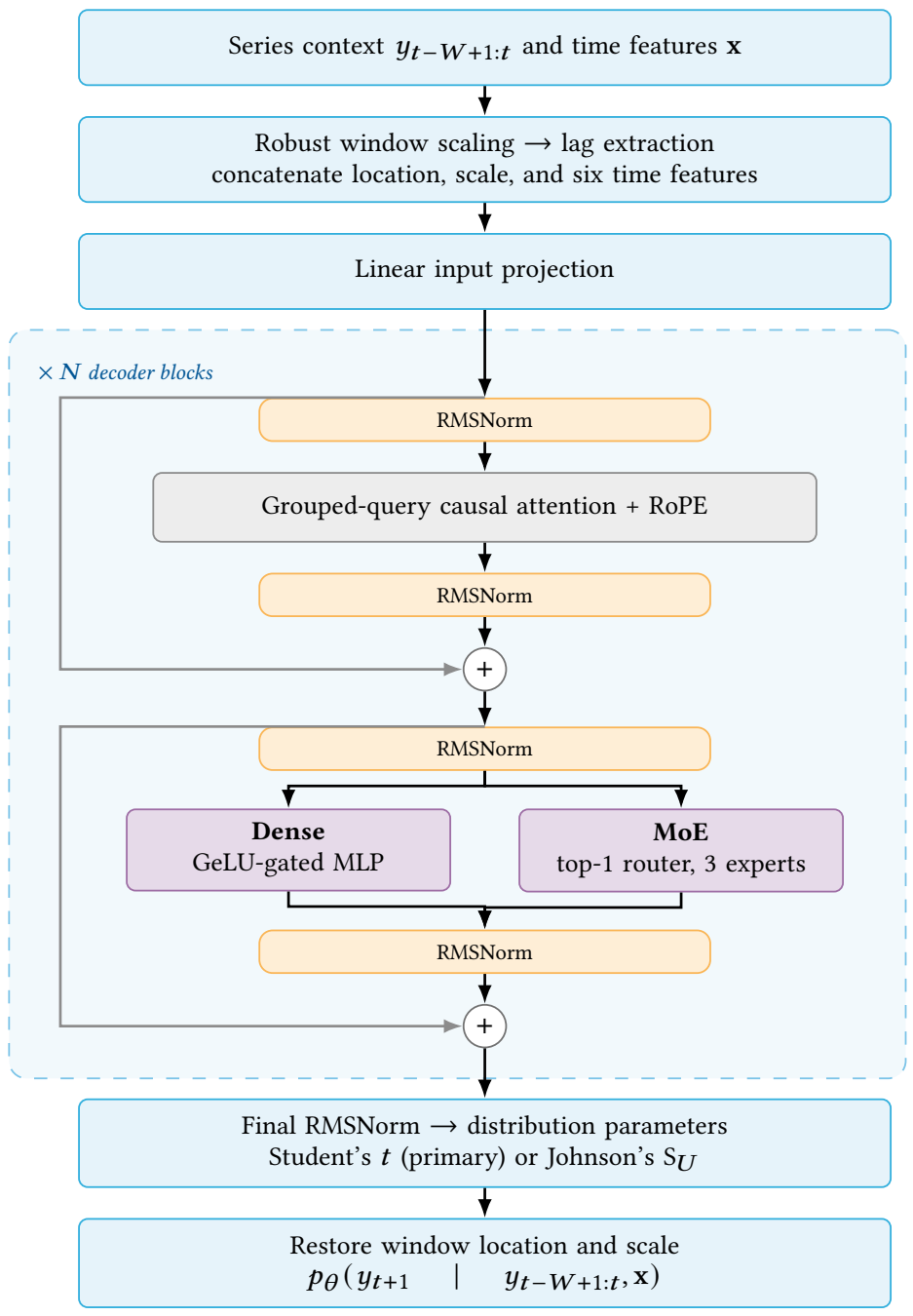

%%%%%%%%%%%%%%%%%%%%%%%%%%%%%%%%%%%%%%%%%%%%%
\section{TailDART TSFM}
\subsection{Forecasting Objective}
For a univariate time series $y_{1:t}$ and a context window of length $W$, the model estimates a one-step conditional distribution
\[
  p_\theta(y_{t+1}\mid y_{t-W+1:t}, \mathbf{x}_{t-W+1:t+1}),
\]
where $\mathbf{x}$ denotes optional time features. The model therefore outputs a full predictive distribution for each forecast step. For evaluation with point-based metrics such as root mean squared error, this distribution is converted into a single forecast by taking its median, which serves as the central predicted value.

\subsection{Architecture}
Figure~\ref{fig:architecture} shows the complete forecasting pipeline.
\paragraph{Input Construction.}
Each input window is first normalized with a configurable scaler, viz.\ mean, standard, robust, or no scaling. The normalized history is then converted into lagged features and augmented with static location and scale summaries. When enabled, six calendar covariates are appended: second-of-minute, minute-of-hour, hour-of-day, day-of-week, day-of-month, and day-of-year. The resulting feature vector is projected through a linear embedding layer into the model hidden space.

\paragraph{Decoder-only Backbone.}
The core model is a decoder-only Transformer. It consists of a stack of repeated blocks, each operating on the embedded lag features and producing a hidden representation for probabilistic forecasting. The backbone follows the usual autoregressive forecasting setup: it processes the input sequence causally and can reuse key-value states during inference for efficient step-by-step prediction. 

\paragraph{Attention Mechanism.}
Inside each block, attention is implemented as causal self-attention with rotary positional embeddings (RoPE) and residual connections. The model uses separate query, key, and value projections, and supports key-value caching for autoregressive decoding. A notable difference from the standard multi-head attention layer is that this implementation uses grouped-query attention (GQA): queries are projected with the number of heads, while keys and values use fewer key and value heads and are then repeated to match the query heads. 

\paragraph{Dense: Feed-forward Sublayer.}
After attention, each block applies a gated feed-forward network. In the \textbf{dense} configuration, the feed-forward module is a  Gaussian error linear unit (GeLU)-gated multilayer perceptron (MLP) with a down-projection back to the hidden dimension. 
The block uses pre-normalization and post-sublayer residual paths, with Root Mean Square Normalization (RMSNorm) applied around both the attention and feed-forward sublayers. This provides the backbone a stable decoder-only structure while keeping the implementation lightweight. 

\paragraph{Mixture-of-Experts}
The architecture also supports an optional sparse variant~\cite{fedus2022switch}. In this mode, the dense MLP is replaced by a top-1 mixture-of-experts (\textbf{MoE}) layer with three experts by default. A router assigns each token to one expert, and an auxiliary load-balancing loss with weight $\lambda_{\mathrm{aux}}=0.01$ (Eq.~(\ref{eq:loss})) encourages more balanced expert usage. 
%This sparse version is treated as an ablation rather than the primary forecasting model. 

\paragraph{Forecast Head.}
The final hidden representation produced by the decoder-only backbone is not mapped directly to a point forecast. Instead, it is passed through a distributional output head, which projects each hidden state into the parameters of a predictive probability distribution. Next, this projection is handled by a learned parameter layer, and the resulting unconstrained outputs are transformed into valid distribution parameters through the corresponding domain map. Depending on the selected head, the model can parameterize \textbf{Student's $t$}, or \textbf{Johnson's} $\textbf{S}_U$, allowing the forecast distribution to capture asymmetry, heavy tails, and multimodality.

The mixture-head variants further extend this distributed-prediction approach by replacing a single predictive family with a finite mixture distribution. Given decoder state $h_t$, the head produces mixture logits $\alpha_k(h_t)$ together with component parameters $\phi_k(h_t)$, thereby defining
\[
  p_\theta(y_{t+1}\mid h_t)
  =
  \sum_{k=1}^{K}
  \pi_k(h_t)\,p_k\!\left(y_{t+1};\phi_k(h_t)\right),
\]
with mixture weights
\[
  \pi_k(h_t)=
  \frac{\exp \alpha_k(h_t)}
       {\sum_{j=1}^{K}\exp \alpha_j(h_t)}.
\]
In this implementation, the mixture-head can either use Student's $t$ components for all $p_k$ or a combination of Student's $t$ and Johnson's $\text{S}_U$ components. 
This implementation selection provides the predictive distribution with greater flexibility in tail behavior and skewness. 
Motivated by the specialization principle of MoE architectures, these mixture distribution heads were developed. Preliminary experiments exhibited unstable optimization and no consistent performance benefit; consequently, these heads are not used in the reported experiments.

In both Dense and MoE variants of the TailDART architecture, however, the probabilistic forecast is still summarized by its median for point-based evaluation. 

\paragraph{Scale Restoration.}
To ensure that forecasts remain interpretable on the original data scale, the input series is normalized before being passed to the backbone, and the corresponding normalization statistics are carried forward through the model. Concretely, the preprocessing step returns the location and scale used for normalization, and these quantities are later used to restore the forecast to the original units. This lets the model learn in a numerically stable, standardized space while still producing predictions that can be interpreted and evaluated on the raw-series scale.

\begin{table}[H]
  \caption{Base and continued-pretraining datasets} % used for the scaling-law analysis.}
  \label{tab:pretraining-datasets}
  \centering
  \scriptsize
  \setlength{\tabcolsep}{3pt}
  \begin{tabularx}{\columnwidth}{@{}l>{\raggedright\arraybackslash}X@{}}
    \toprule
    Source/group & Dataset names \\
    \midrule
    ETT &
      \path{ETTh1}, \path{ETTh2}, \path{ETTm1}, \path{ETTm2} \\
    GluonTS electricity/energy &
      \path{australian_electricity_demand}, \path{electricity_hourly},
      \path{london_smart_meters_without_missing}, \path{solar_10_minutes},
      \path{wind_farms_without_missing} \\
    GluonTS transport/mobility &
      \path{pedestrian_counts}, \path{uber_tlc_hourly}, \path{traffic},
      \path{traffic_nips} \\
    GluonTS weather/environment &
      \path{kdd_cup_2018_without_missing}, \path{saugeenday},
      \path{sunspot_without_missing}, \path{weather} \\
    GluonTS exchange rates &
      \path{exchange_rate} \\
    GluonTS M4 &
      \path{m4_daily}, \path{m4_weekly}, \path{m4_monthly} \\
    Short-history additions &
      \path{fred_md}, \path{hospital}, \path{tourism_monthly},
      \path{m4_hourly}, \path{nn5_daily_without_missing}, \path{nn5_weekly} \\
    \bottomrule
  \end{tabularx}
\end{table}
%%%%%%%%%%%%%%%%%%%%%%%%%%%%%%%%%%
\subsection{Base Pretraining}
\paragraph{Dataset.} The model was pretrained on a twenty-one heterogeneous collection of publicly available time-series datasets (Table~\ref{tab:pretraining-datasets}). The pretraining pool includes the ETT benchmark series~\cite{haoyietal-informer-2021}, electricity and energy datasets obtained through the GluonTS dataset repository\footnote{\url{https://ts.gluon.ai/stable/api/gluonts/gluonts.dataset.repository.datasets.html}}~\cite{alexandrov2020gluonts}, transport and mobility datasets, weather and environment series, the exchange-rate dataset, and the daily, weekly, and monthly subsets of M4. This mixture was intended to expose the model to a broad range of temporal patterns, including short-term dynamics, strong seasonality, long-range dependence, and differing levels of noise and persistence. The resulting base checkpoints are used for the scaling-law analysis.

% DID WE ALSO DO THE FOLLOWING?
%Stratified Domain Sampling for LagLlama:
%Because datasets in public repositories vary from thousands of series to just a few dozen, naive batch sampling causes small datasets to be ignored or large %datasets to overfit. Lag-Llama employs stratified sampling across datasets, ensuring each domain is sampled proportionally to its variance and representation.

\paragraph{Data Augmentation.}
To improve robustness, the pretraining pipeline applies data augmentation with probability 0.5, implying that, for about half of the training examples, the model is fed a perturbed version of the original series rather than the raw input. The augmentation pool includes random time-domain perturbations and frequency-domain transformations, designed to reduce sensitivity to local noise, temporal distortions, and spectral variation. In particular, frequency masking is applied with rate 0.5, and frequency mixing is applied with rate 0.25. 

\paragraph{Inference and Loss.}
At inference time, the model takes a normalized lagged context and optional time features, then maps them to the parameters of a predictive distribution. Autoregressive decoding is supported through key-value caching, which reuses previous attention states during step-by-step generation. Multiple sample paths can then be drawn from the predictive distribution, and the resulting forecasts are transformed back to the original scale using the stored location and scale statistics. 

Training uses teacher forcing with shifted future targets, thus each prediction step conditions on previously observed values. The objective is an observed-value-weighted negative log-likelihood (NLL), augmented by an auxiliary load-balancing term for MoE models:
\begin{equation}
  \mathcal{L} = -\frac{\sum_t m_t\log p_\theta(y_t)}{\sum_t m_t}
  + \lambda_{\mathrm{aux}}\mathcal{L}_{\mathrm{MoE}},
  \label{eq:loss}
\end{equation}
where $m_t\in\{0,1\}$ masks unobserved values and $\lambda_{\mathrm{aux}}$ weights the auxiliary load-balancing loss $\mathcal{L}_{\mathrm{MoE}}$. The auxiliary term is zero for Dense models (Figure~\ref{fig:architecture}).

\paragraph{Implementation Details.}
All pretraining configurations use AdamW with a batch size of 512, a context length of 32, a prediction length of 1, robust scaling, time features, no dropout, series weighting, and 1 validation pass per epoch. Hyperparameters are selected by validation loss. 

For the Student's $t$ head, the learning rate is $5{\times}10^{-4}$, with weight decay set to $10^{-3}$ for the dense model and $10^{-4}$ for the MoE model. For the Johnson's $\text{S}_U$ head, the learning rate is $10^{-3}$ with the same architecture-specific weight decay settings. 

\subsection{Short-History Continued Pretraining}
To improve robustness to incomplete histories, each base checkpoint undergoes a second training stage using the same forecasting objective. This stage retains the original 21-dataset pool and adds six datasets from the GluonTS dataset repository: \path{fred_md}, \path{hospital}, \path{tourism_monthly}, \path{m4_hourly}, \path{nn5_daily_without_missing}, and \path{nn5_weekly}. Series shorter than the required history length are retained by left-padding them with masked values. The original and additional dataset pools each receive a 50\% sampling probability, with series weighting applied within each pool. Stochastic left-padded history augmentation is applied with probability 0.15.

For the Student’s $t$ setting, checkpoints are selected by the standard validation NLL, whereas the Johnson's $\text{S}_U$ setting additionally averages validation NLL over retained-history ratios from 0.9 to 1.0.

%%%%%%%%%%%%%%%%%%%%%%%%%%%%%%%%%%%%%%%%%%%%%%%%%%
\FloatBarrier
\begin{figure}[t]
  \centering
  \par\vspace{2pt}
  \begin{subfigure}{\columnwidth}
    \centering
    \includegraphics[width=0.83\linewidth]{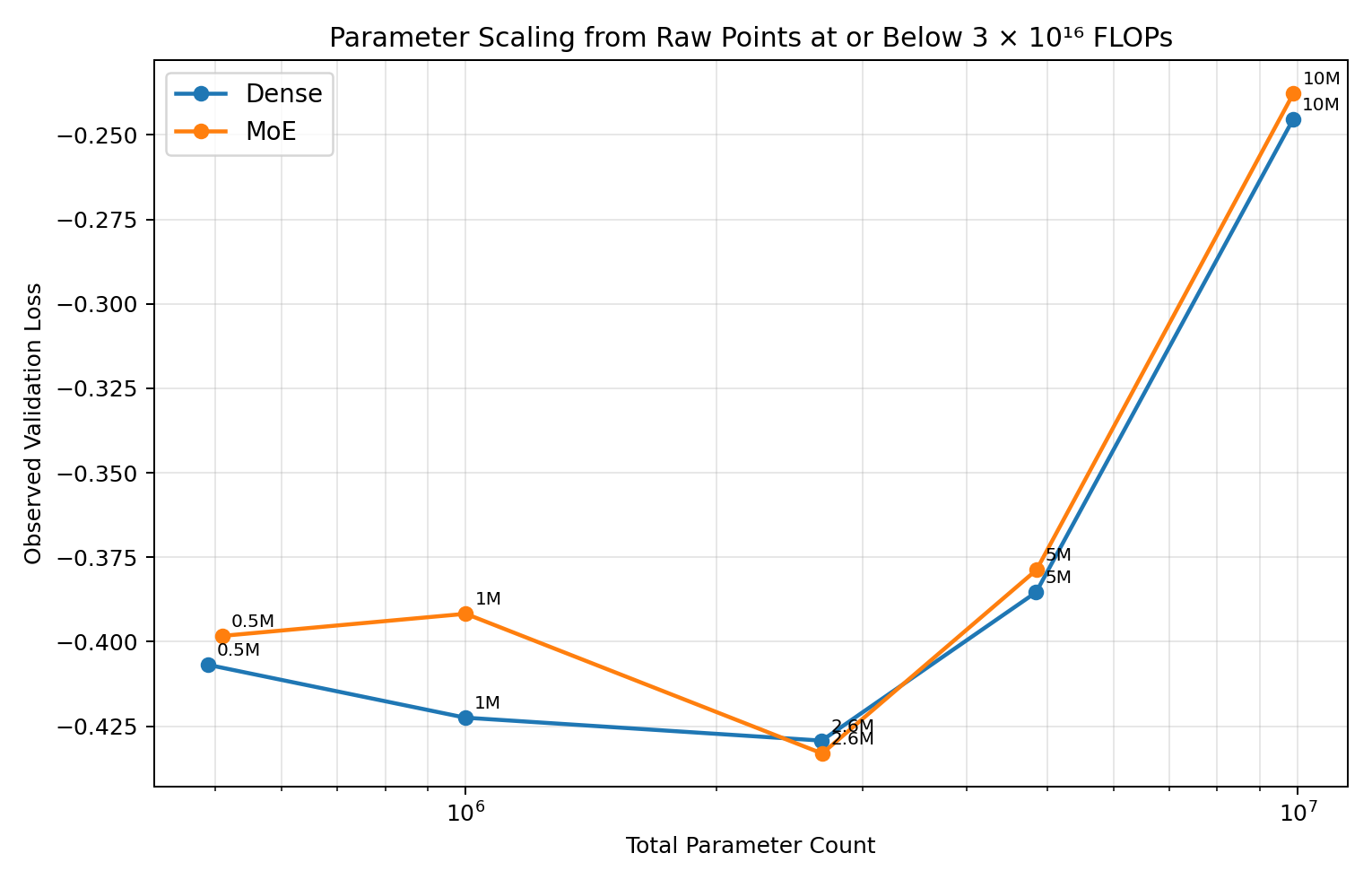}
    \caption{Parameter scaling at fixed compute.}
    \label{fig:scaling-parameters}
  \end{subfigure}
  \begin{subfigure}{\columnwidth}
    \centering
    \includegraphics[width=0.83\linewidth]{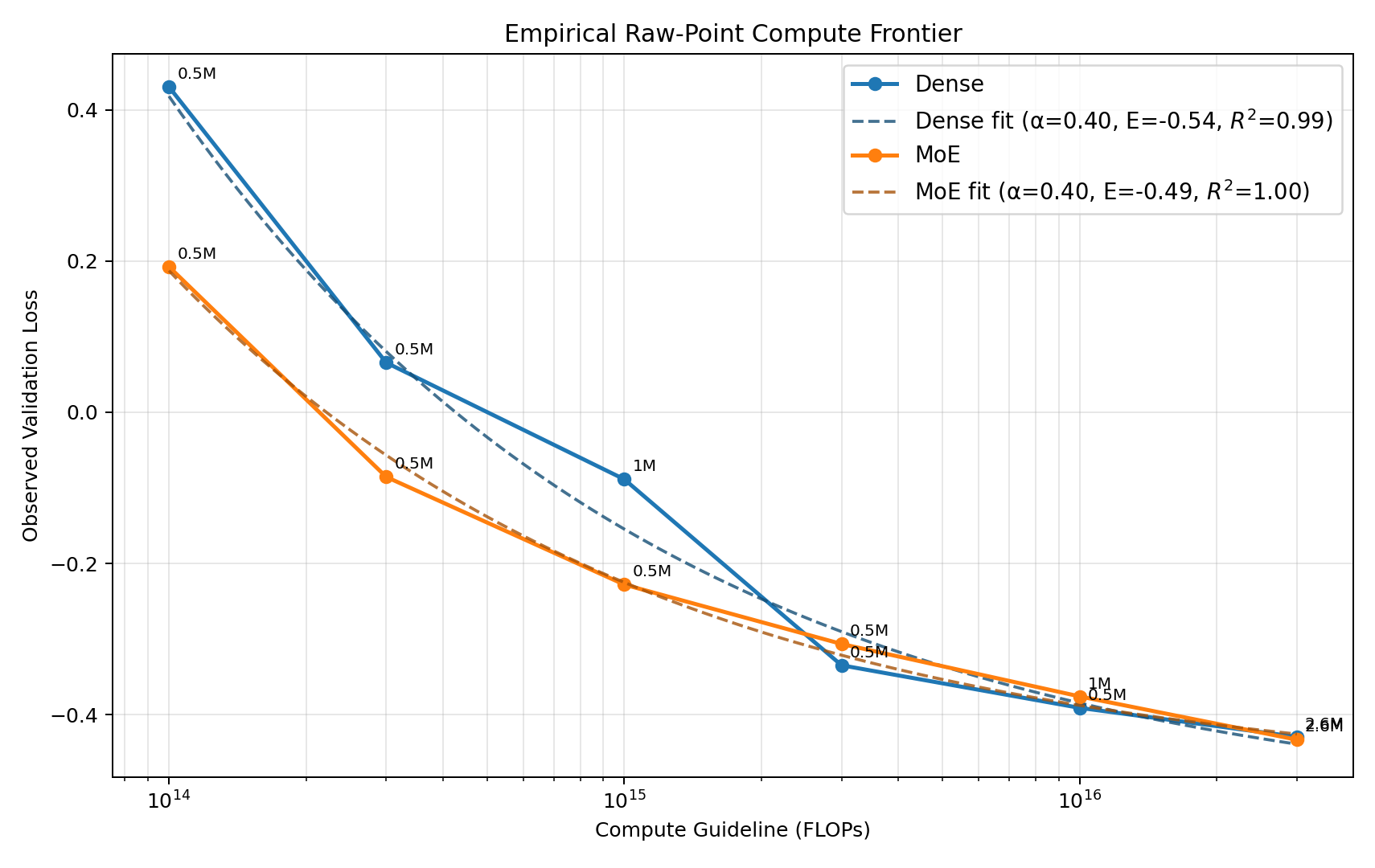}
    \caption{Compute-efficient frontier.}
    \label{fig:scaling-compute-frontier}
  \end{subfigure}
  \par\vspace{2pt}
  \begin{subfigure}{\columnwidth}
    \centering
    \includegraphics[width=0.83\linewidth]{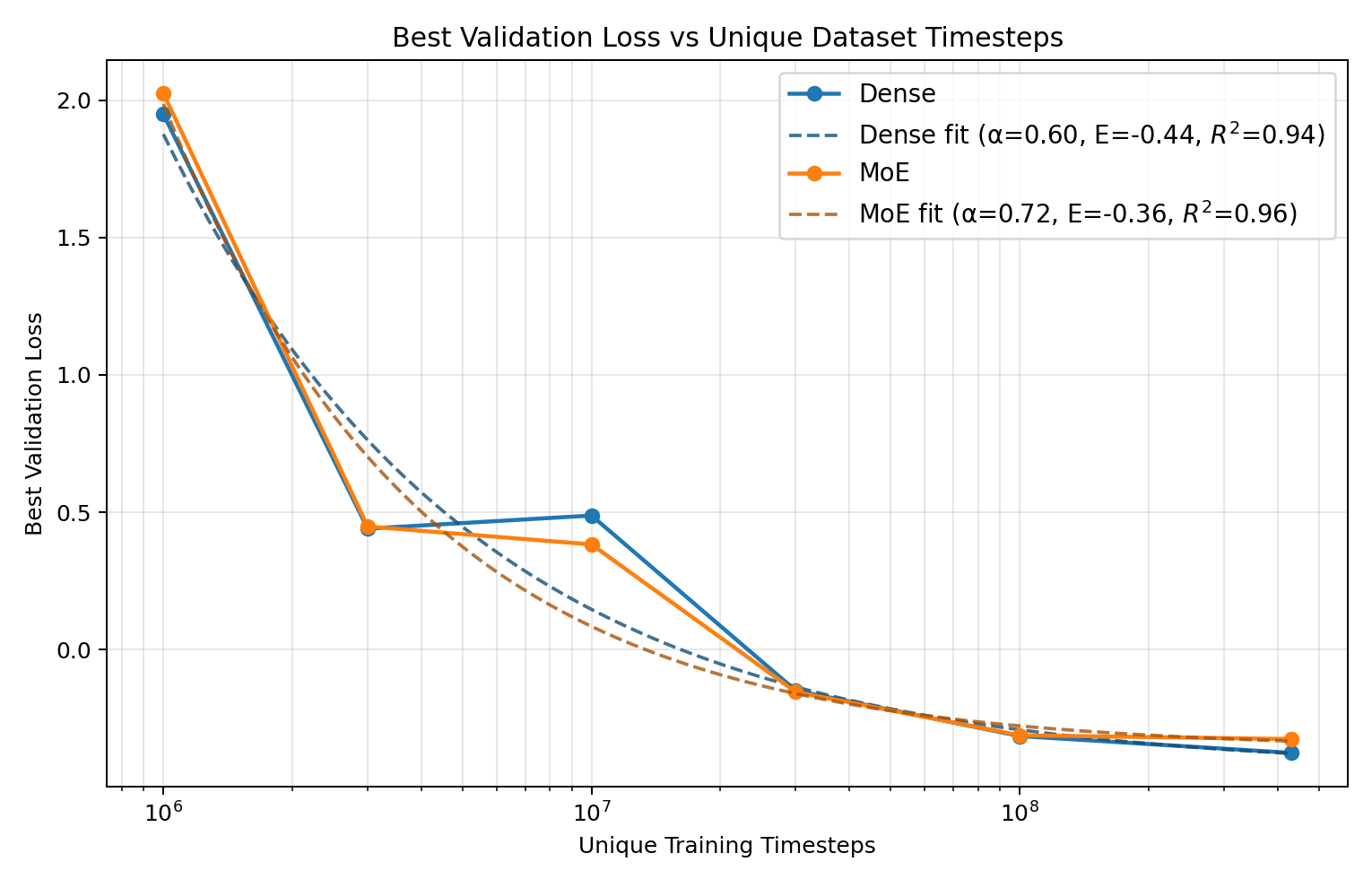}
    \caption{Dataset-size scaling for 2.6M models.}
    \label{fig:scaling-dataset}
  \end{subfigure}
  %\Description{Three panels summarize the TailDART scaling sweeps for dense and MoE variants: (a) validation loss versus estimated FLOPs with the compute-efficient frontier, (b) validation loss versus total parameters at fixed compute, and (c) best validation loss ($\mathcal{L}$) versus the number of training timesteps.}
  \caption{Scaling behavior of dense and MoE \modelname{}. Lower validation NLL is better in all panels.}
  \label{fig:scaling}
\end{figure}

\begin{table*}[!htp]
  \caption{Performance of TailDART against baselines.}  
  \label{tab:main}
  \centering
  \small
  \begin{tabular}{lrrrrrrrrrr}
    \toprule
    & & \multicolumn{3}{c}{Daily} & \multicolumn{3}{c}{Weekly} & \multicolumn{3}{c}{Monthly} \\
    \cmidrule(lr){3-5}\cmidrule(lr){6-8}\cmidrule(lr){9-11}
    Model & Params (M) & MASE & RMSE & Latency (ms) & MASE & RMSE & Latency (ms) & MASE & RMSE & Latency (ms) \\
    \midrule
    \modelname{} (Dense) & 2.68 & 0.859 & 0.0201 & \textbf{0.668} & 0.899 & 0.0411 & \textbf{0.689} & 0.842 & 0.0977 & \textbf{0.912} \\
    \modelname{} (MoE)   & 2.68 & 0.865 & 0.0204 & 1.058 & 0.908 & 0.0421 & 1.094 & 0.848 & 0.0995 & 1.629 \\
    \midrule
    Lag-Llama  & 2.45 & 0.854 & 0.0204 & 1.288 & 0.906 & 0.0423 & 0.986 & 0.855 & 0.1002 & 1.234 \\
    Chronos-T5-tiny    & 8.39 & 0.819 & 0.0191 & 1.047 & 0.857 & \textbf{0.0384} & 0.704 & 0.821 & 0.0931 & 1.021 \\
    Sundial-base    & 128.00 & \textbf{0.798} & \textbf{0.0185} & 7.019 & \textbf{0.846} & 0.0385 & 6.907 & \textbf{0.802} & \textbf{0.0929} & 8.042 \\
    Moirai     & 13.80 & 0.867 & 0.0201 & 10.971 & 0.904 & 0.0432 & 11.431 & 0.853 & 0.1000 & 16.806 \\
    Moirai-MoE & 117.00 & 0.825 & 0.0190 & 3.414 & 0.876 & 0.0396 & 3.010 & 0.819 & 0.0949 & 4.451 \\
    \bottomrule
  \end{tabular}
\end{table*}

%%%%%%%%%%%%%%%%%%%%%%%%%%%%%%%%%%%%%%%%%%%%%%%%%%%%%%%%%%%%%
\section{Empirical Scaling Analysis}
\label{sec:scaling_laws}

To evaluate model scalability under computational constraints, an empirical scaling study, adapted from the framework of Kaplan et al.~\cite{kaplan2020scaling}, was conducted for both Dense and MoE variants of TailDART. The analysis investigates validation loss ($\mathcal{L}$) trajectory as a function of training compute ($C$), total active non-embedding parameter count ($N$), and unique dataset timesteps ($D$).

To align with standard scaling-law formulations, dataset volume $D$ is defined as the total number of unique, non-overlapping temporal observations (timesteps) across all training series before context window extraction. Training compute is estimated as $C \approx 6NT$, where $T$ counts processed training tokens, including repeated observations, and is distinct from the unique dataset volume $D$.

\paragraph{Parameter Scaling and Empirical Size Selection}
Figure~\ref{fig:scaling-parameters} illustrates the relationship between $\mathcal{L}$ and $N$ under a fixed compute ceiling ($C \le 3 \times 10^{16}\,\text{FLOPs}$). Rather than a monotonic reduction in loss, parameter scaling traces a distinct U-shaped trajectory:
\begin{itemize}
    \item $\mathcal{L}$ decreases from $0.5\,\text{M}$ parameters to a minimum at $N^* = 2.6\,\text{M}$ parameters (loss $\approx -0.43$).
    \item Beyond $2.6\,\text{M}$ parameters, validation performance degrades markedly, rising to $-0.24$ at $10\,\text{M}$ parameters.
\end{itemize}
Under a fixed compute budget $C \approx 6NT$, increasing active parameter count $N$ reduces processed training tokens $T \approx C/(6N)$ and optimizer updates. The higher losses of the $5\,\text{M}$ and $10\,\text{M}$ configurations may therefore reflect insufficient optimization rather than an intrinsic capacity limit.
%This empirical optimum justifies the selection of $2.68\,\text{M}$ parameters as the core architecture size for \modelname{}.

\paragraph{Compute-Efficient Frontier Dynamics}
Figure~\ref{fig:scaling-compute-frontier} presents the empirical compute-efficient frontier evaluated across $2.5$ orders of magnitude: $10^{14}$ to $3 \times 10^{16}\,\text{FLOPs}$. When fitted to a power-law trend $\mathcal{L}(C) = (C_c / C)^\alpha + E$, both Dense ($R^2 = 0.99$) and MoE ($R^2 = 1.00$) architectures exhibit power exponents of $\alpha = 0.40$. The frontier reveals two distinct operational regimes:
\begin{itemize}
    \item Low-Compute Regime ($C \le 10^{15}\,\text{FLOPs}$): The MoE architecture demonstrates superior efficiency, attaining significantly lower validation loss at $10^{14}\,\text{FLOPs}$ ($0.19$ vs.\ $0.43$ for Dense). Sparse gating allows the model to expand capacity without increasing per-step floating-point operations.
    \item High-Compute Regime ($C > 3 \times 10^{15}\,\text{FLOPs}$): As training compute increases, the performance gap narrows. At $3 \times 10^{16}\,\text{FLOPs}$, the Dense architecture extrapolates toward a slightly lower error floor than MoE. %($E = -0.54$ vs.\ $-0.49$ for \textsf{MoE}).
\end{itemize}

\paragraph{Dataset Scaling and Token Saturation}
Figure~\ref{fig:scaling-dataset} evaluates validation loss $\mathcal{L}(D)$ across unique dataset timesteps ranging from $10^6$ to $5 \times 10^8$. Fitted to $\mathcal{L}(D) = (D_c / D)^\alpha + E$, both variants demonstrate strong alignment ($R^2 = 0.94$ for Dense, $R^2 = 0.96$ for MoE). Although MoE displays a steeper initial scaling exponent ($\alpha = 0.72$ vs.\ $\alpha = 0.60$ for Dense) in data-constrained regimes ($10^6$--$10^7$ timesteps), returns diminish beyond $3 \times 10^7$ timesteps. At the upper data limit ($5 \times 10^8$ timesteps), Dense settles at a lower irreducible loss parameter ($E = -0.44$) compared to MoE ($E = -0.36$).

Collectively, these empirical scaling trajectories provide a principled, data-driven rationale for the model architecture design. The parameter scaling curve motivates $2.68\,\text{M}$ parameters ($N^*$) as an empirical design choice under the tested training setup, rather than a robust capacity bound.
This size provides a practical balance between validation performance and training cost in these experiments.

While these findings offer clear design guidelines, the empirical scope remains subject to notable limitations: the localized parameter range ($0.5\,\text{M}$--$10\,\text{M}$) and six-budget sampling points are not sufficient to establish universality of the power laws across the multi-order-of-magnitude range, as shown in~\cite{kaplan2020scaling}. All scaling runs use the same AdamW learning rate ($10^{-3}$) and weight decay ($10^{-4}$), without size-specific tuning. Model size is therefore confounded with optimization quality, and better-tuned larger models could change the observed $N^*$.

%, and active-parameter FLOP proxies for \textsf{MoE} omit routing and communication overheads. 

Nonetheless, the empirical frontier indicates that dense models achieve superior Pareto efficiency compared to the larger configurations under the tested training setup.

%%%%%%%%%%%%%%%%%%%%%%%%%%%%%%%%%%%%%%%%%%%%%%%%%%%%%%%
\section{Experimental Setups: Zero-Shot Inference}
\paragraph{Datasets}
For zero-shot evaluation, the study uses BForTFin~\cite{cheong2025bfortfin}, a finance-domain benchmark containing 73 time series across eight asset classes: Foreign Exchange (\textbf{FX}) (8), Emerging Market FX (\textbf{EM FX} (14), \textbf{MSCI} equity indices (19), Fixed Income (\textbf{Rates}) (10), \textbf{REITs} (5), Individual \textbf{Equities} (6), \textbf{ETFs} (5), and Cryptocurrency (\textbf{Crypto}) (6).
Each series is converted to three sampling frequencies: \textbf{Daily} (1-day), \textbf{Weekly} (5-day), and \textbf{Monthly} (20-day), and then transformed into simple arithmetic returns, $r_t=P_t/P_{t-1}-1$, computed between consecutive time points.

To examine the effect of context length on point forecasting, the return series are further reformatted into supervised samples with fixed look-back windows $L$. For daily data, $L \in \{5, 10, 20, 60, 126, 252\}$; for weekly data, $L \in \{4, 9, 13, 26, 52, 104\}$; and for monthly data, $L \in \{6, 12, 18, 24, 60\}$. This protocol yields 66{,}088 daily, 16{,}944 weekly, and 2{,}833 monthly forecast instances, for a total of 85{,}865 zero-shot forecasts.

\paragraph{Zero-shot protocol and contamination check.}
Here, zero-shot denotes frozen-parameter inference without BForTFin training. The saved dataset lists were checked for both reported models: BForTFin is excluded from base and continued pretraining, including their validation pools. However, instrument-level matching and duplicate checks against FX and Rates were not completed, so source-level overlap cannot be ruled out, although no such overlap has been confirmed. The local BForTFin~\cite{cheong2025bfortfin} source files span 2015-01-01 to 2025-07-07. In the loaded training partitions, \path{exchange_rate} has generated business-day indices from 1990-01-01 to 2013-04-05, while \path{fred_md}, added during continued pretraining, spans January 1959 to July 2018. The M4 subsets use placeholder dates beginning on 1750-01-01 in the GluonTS representation rather than their original calendar dates.

\paragraph{TailDART for Benchmarking.}
Two pretrained variants of \modelname: Dense-2.68M and MoE-2.68M with Johnson's $\text{S}_U$ distribution head are used for this benchmarking study.
Both checkpoints underwent continued pretraining for short-history robustness using the datasets listed in Table~\ref{tab:pretraining-datasets}.

%%%%%%%%%%%%%%%%
\paragraph{Baseline TSFMs}
The zero-shot comparison includes Lag-Llama~\cite{rasul2023lagllama}, Chronos-T5-tiny~\cite{ansari2024chronos}, Sundial-base-128M~\cite{liu2025sundial}, Moirai~\cite{woo2024moirai}, and Moirai-MoE~\cite{liu2025moiraimoe}. Among these, Moirai 1.1-R-small and Moirai-MoE 1.0-R-small are evaluated because their continuous distributional outputs are directly comparable to the probabilistic heads used by \modelname{}.%
\footnote{Moirai-MoE 1.0-R-small has 117M total parameters and activates 11M parameters per token, routing to 2 of 32 experts per layer~\cite{liu2025moiraimoe}. Total parameter counts are reported for all models; the ``0.1B'' label on the public checkpoint is a rounded total count rather than the activated count.}

%The sparse \modelname{} variant is reported separately as an ablation and is not included in the main baseline set.

%%%%%%%%%%%%%%%%%%
\paragraph{Compute.}
Tests are conducted on a RunPod instance equipped with one NVIDIA GeForce RTX 4090 GPU, 16 AMD Ryzen 9 7950X vCPUs, and 61 GB of memory. 

%%%%%%%%%%%%%%%%%
\paragraph{Evaluation Metrics} 
Model accuracy is evaluated primarily using Mean Absolute Scaled Error (\textbf{MASE}), with Root Mean Squared Error (\textbf{RMSE}) serving as a secondary metric. For a given look-back window $L$, the absolute error of the one-step-ahead forecast is normalized by the mean absolute first difference of the preceding historical sequence of length $L$, which corresponds to the baseline error of a one-step naive persistence forecast. These ratios are subsequently averaged across all look-back windows $L$ per series. This normalization standardizes performance across series with varying return scales, where $\text{MASE} < 1$ indicates superior performance relative to the naive baseline, and $\text{MASE} > 1$ denotes inferior performance. 

For the frequency-level values reported in the comparison tables, MASE and RMSE are macro-averaged over the available asset-group--look-back-window cells. In particular, for metric $q\in\{\mathrm{MASE},\mathrm{RMSE}\}$ and frequency $f$, the reported value is
\begin{equation}
    \bar q_f = \frac{1}{|\mathcal{A}|\,|\mathcal{L}_f|}
    \sum_{a\in\mathcal{A}}\sum_{L\in\mathcal{L}_f}q_{a,L},
\end{equation}
where $\mathcal{A}$ is the set of eight reporting groups and $\mathcal{L}_f$ is the set of look-back windows for frequency $f$. Thus, each asset-group--window cell receives equal weight rather than being weighted by its number of forecasts. Reported latency uses forecast-count weighting instead: if timing record $i$ contains elapsed time $t_i$ seconds for $n_i$ forecasts, then
\begin{equation}
    \operatorname{Latency}_f = 10^3\frac{\sum_{i\in f}t_i}{\sum_{i\in f}n_i}\ \text{ms per forecast}.
\end{equation}

Inference efficiency is evaluated via \textbf{latency}, defined as the total wall-clock time divided by the number of generated forecasts. To ensure fair benchmarking, latency is compared exclusively among models with equivalent one-step output workloads. All timing measurements are conducted on a single RunPod instance using identical hardware, batch sizes, and CUDA synchronization protocols.\footnote{The Lag-Llama estimator does not natively propagate its single-pass sampling flag to the inference Lightning module. To prevent the model from defaulting to 100 parallel samples---which substantially increases runtime and memory overhead---this flag is set explicitly on the module during evaluation.}

Consequently, lower values indicate better performance across all three metrics: MASE, RMSE, and Latency.

%%%%%%%%%%%%%%%%%%%%%%%%%%%%%%%%%%%%%%%%%%%%%%%%%%%%%%
\section{Results: Zero-Shot Experiments}

\subsection{Aggregated Performance}
\label{sec:results}
Table~\ref{tab:main} presents the comparative evaluation of \modelname{} against state-of-the-art TSFMs across Daily, Weekly, and Monthly forecasting horizons. %Performance is evaluated in terms of forecast accuracy (\textbf{MASE} and \textbf{RMSE}) and computational efficiency (\textbf{Latency} in milliseconds).

\paragraph{Inference Efficiency and Scale Efficiency}
\modelname{} (Dense) achieves the lowest inference latency across all evaluated temporal frequencies, requiring only $0.668\,\text{ms}$ for Daily, $0.689\,\text{ms}$ for Weekly, and $0.912\,\text{ms}$ for Monthly forecasts. Compared to Lag-Llama ($2.45\,\text{M}$ parameters), \modelname{} (Dense) yields a latency reduction of up to $48.1\%$ ($0.668\,\text{ms}$ vs.\ $1.288\,\text{ms}$ on Daily data) while simultaneously improving predictive accuracy on Weekly ($\text{MASE } 0.899$ vs.\ $0.906$) and Monthly ($\text{MASE } 0.842$ vs.\ $0.855$) horizons. Furthermore, \modelname{} (Dense) consistently outperforms Moirai ($13.80\,\text{M}$ parameters) across all three frequency domains in both MASE and RMSE, despite utilizing over $5\times$ fewer parameters and operating up to $18.4\times$ faster ($0.912\,\text{ms}$ vs.\ $16.806\,\text{ms}$ on Monthly forecasts).

\paragraph{Pareto Accuracy--Latency Trade-offs}
While larger foundation baselines such as Sundial-base-128M ($128.00\,\text{M}$ parameters) and Chronos-T5-tiny ($8.39\,\text{M}$ parameters) attain lower absolute error metrics (e.g., Sundial reaches a Daily MASE of $0.798$), they demand substantially greater computational resources. Specifically, Sundial requires $47.8\times$ more parameters and incurs an inference latency penalty between $8.8\times$ and $10.5\times$ relative to \modelname{} (Dense). Consequently, \modelname{} defines a favorable Pareto boundary for resource-constrained or real-time deployment settings where microsecond-level latency constraints preclude the use of hundred-million-parameter architectures.

\paragraph{Dense vs.\ Mixture-of-Experts Variant}
Comparing the two internal variants of \modelname{} reveals that the Dense architecture strictly dominates the MoE implementation at the $2.68\,\text{M}$ parameter scale. The MoE variant exhibits higher error metrics across all horizons (e.g., Daily MASE of $0.865$ vs.\ $0.859$) and introduces a $58.4\%$ to $78.6\%$ latency overhead ($1.058\,\text{ms}$--$1.629\,\text{ms}$). This performance drop indicates that the gating mechanism and sparse routing overhead might introduce unnecessary computational latency without providing sufficient capacity gains at this dense model scale.

\begin{table}[H]
  \caption{Asset-level comparison of TailDART and Lag-Llama. $\Delta$ is the percentage change from Lag-Llama to \modelname{}.}
  \label{tab:findart-lagllama-asset-groups}
  \centering
  \scriptsize
  \setlength{\tabcolsep}{2pt}
  \renewcommand{\arraystretch}{0.92}
  \begin{tabular*}{\columnwidth}{@{}c@{\hspace{1.5pt}}l@{\hspace{3pt}}!{\vrule width 0.3pt}@{\hspace{3pt}\extracolsep{\fill}}rrrrrr@{}}
    \toprule
    & & \multicolumn{3}{c}{MASE} & \multicolumn{3}{c}{RMSE} \\
    \cmidrule(lr){3-5}\cmidrule(lr){6-8}
    & Asset group & \modelname{} & Lag-Llama & $\Delta$ & \modelname{} & Lag-Llama & $\Delta$ \\
    \midrule
    \multirow{8}{*}{\rotatebox[origin=c]{90}{Daily}}
    & FX     & 0.827 & 0.823 & +0.5\% & 0.0060 & 0.0058 & +2.4\% \\
    & EM FX     & 0.847 & 0.819 & +3.3\% & 0.0047 & 0.0044 & +7.0\% \\
    & MSCI      & 0.864 & 0.869 & -0.5\% & 0.0164 & 0.0165 & -0.4\% \\
    & Rates     & 0.847 & 0.839 & +1.0\% & 0.0246 & 0.0243 & +1.3\% \\
    & REITs     & 0.955 & 0.931 & +2.6\% & 0.0258 & 0.0227 & +13.4\% \\
    & Equities  & 0.845 & 0.862 & -2.0\% & 0.0144 & 0.0145 & -0.4\% \\
    & ETFs      & 0.870 & 0.866 & +0.5\% & 0.0168 & 0.0163 & +2.5\% \\
    & Crypto    & 0.815 & 0.824 & -1.1\% & 0.0520 & 0.0585 & -11.1\% \\
    \midrule
    \multirow{8}{*}{\rotatebox[origin=c]{90}{Weekly}}
    & FX     & 0.898 & 0.902 & -0.5\% & 0.0133 & 0.0133 & +0.3\% \\
    & EM FX     & 0.941 & 0.938 & +0.3\% & 0.0101 & 0.0102 & -1.1\% \\
    & MSCI      & 0.895 & 0.903 & -0.8\% & 0.0324 & 0.0327 & -1.1\% \\
    & Rates     & 0.861 & 0.854 & +0.8\% & 0.0496 & 0.0477 & +4.0\% \\
    & REITs     & 0.964 & 0.994 & -3.0\% & 0.0414 & 0.0423 & -2.0\% \\
    & Equities  & 0.866 & 0.866 & +0.0\% & 0.0305 & 0.0306 & -0.1\% \\
    & ETFs      & 0.907 & 0.914 & -0.7\% & 0.0324 & 0.0330 & -1.8\% \\
    & Crypto    & 0.857 & 0.881 & -2.8\% & 0.1189 & 0.1291 & -7.9\% \\
    \midrule
    \multirow{8}{*}{\rotatebox[origin=c]{90}{Monthly}}
    & FX     & 0.735 & 0.746 & -1.4\% & 0.0235 & 0.0238 & -1.2\% \\
    & EM FX     & 0.799 & 0.803 & -0.6\% & 0.0183 & 0.0186 & -1.8\% \\
    & MSCI      & 0.869 & 0.886 & -1.9\% & 0.0619 & 0.0622 & -0.3\% \\
    & Rates     & 0.955 & 0.937 & +1.9\% & 0.1039 & 0.1000 & +3.9\% \\
    & REITs     & 0.949 & 0.968 & -2.0\% & 0.0984 & 0.0978 & +0.6\% \\
    & Equities  & 0.848 & 0.900 & -5.8\% & 0.0543 & 0.0567 & -4.2\% \\
    & ETFs      & 0.756 & 0.736 & +2.7\% & 0.0563 & 0.0527 & +6.9\% \\
    & Crypto    & 0.825 & 0.861 & -4.2\% & 0.3648 & 0.3897 & -6.4\% \\
    \bottomrule
  \end{tabular*}
\end{table}

%%%%%%%%%%%%%%%%%
\subsection{Asset-Class Level Granular Analysis}
\label{sec:asset_level_analysis}

To establish a direct, controlled baseline comparison against a model of equivalent parameter scale ($\sim 2.5\,\text{M}$ parameters), Table~\ref{tab:findart-lagllama-asset-groups} provides a granular breakdown comparing \modelname{} against Lag-Llama across eight asset classes.
%: FX, EM Fx, MSCI, Rates, REITs, Equities, ETFs, and Crypto.

\paragraph{Horizon-Dependent Generalization}
The asset-level results reveal a clear pattern: the relative advantage of \modelname{} scales monotonically with lower frequency. At the Daily frequency, Lag-Llama retains a slight edge on low-volatility macro series (such as FX and Rates), with \modelname{} outperforming on 3 of 8 asset groups. However, as the horizon expands to Weekly and Monthly forecasts, \modelname{} demonstrates superior generalizability, achieving lower MASE on 5 of 8 asset classes at the Weekly resolution and 6 of 8 asset classes at the Monthly resolution. Notably, on Monthly Equities, \modelname{} yields a $5.8\%$ reduction in MASE ($0.848$ vs.\ $0.900$) and a $4.2\%$ reduction in RMSE ($0.0543$ vs.\ $0.0567$) relative to Lag-Llama.

\paragraph{Robustness in High-Volatility Regimes}
Across all evaluated temporal frequencies, \modelname{} demonstrates marked gains on asset classes characterized by non-Gaussian returns, high kurtosis, and frequent regime shifts:
\begin{itemize}%[leftmargin=*]
    \item Cryptocurrency: \modelname{} outperforms Lag-Llama across all metrics and timeframes, exhibiting substantial reductions in square-error loss ($\Delta\text{RMSE}$ of $-11.1\%$ Daily, $-7.9\%$ Weekly, and $-6.4\%$ Monthly). This underscores \modelname{}'s ability to accommodate heavy-tailed distributions without suffering catastrophic error spikes during market shocks.
    \item Equity-Linked Assets: On MSCI indices and individual Equities, \modelname{} maintains consistent leads across nearly all horizons, achieving peak MASE improvements of $-1.9\%$ and $-5.8\%$ on Monthly data, respectively.
\end{itemize}

\paragraph{Domain Trade-offs}
Conversely, Lag-Llama exhibits strong baseline performance on policy-constrained or highly mean-reverting series, such as Rates (where it achieves a $1.0\%$--$1.9\%$ lower MASE across frequencies) and Daily EM FX ($+3.3\%$ MASE for \modelname{}). In these lower-volatility regimes, the persistent autoregressive bias inherent to Lag-Llama provides effective regularization against high-frequency noise, whereas \modelname{} is better optimized for capturing broader cross-asset temporal dynamics across extended horizons.

%\paragraph{Discussion: Analysis of Latency Efficiency}
Overall, across all evaluated benchmarks, \modelname{} (Dense) demonstrates a pronounced latency advantage, running up to $48.1\%$ faster than Lag-Llama despite operating at an equivalent parameter scale ($\sim 2.5\,\text{M}$ parameters). Trivial explanations for this efficiency gap can be ruled out: \modelname{} carries a slightly larger parameter footprint ($2.68\,\text{M}$ vs.\ $2.45\,\text{M}$), and both models are evaluated under strictly matched workload conditions using an explicit single-pass inference path.

From an architectural perspective, a candidate driver for this speedup is the adoption of GQA for TailDART in place of standard multi-head attention. In small-model, real-time inference regimes where execution is frequently memory-bandwidth bound, GQA reduces key-value projection dimensions and activation memory overhead, theoretically improving operational intensity. 

GQA, however, is presented as a plausible, theoretically grounded hypothesis rather than a proven isolated cause because wall-clock latency aggregates both high-level structural choices and low-level execution artifacts, such as PyTorch SDPA kernel backends, memory layout alignments, and framework overheads. 
%A definitive causal attribution would require a dedicated ablation suite holding non-attention layers, CUDA execution backends, and memory layouts strictly constant.

%%%%%%%%%%%%%%%%%%%%%%%%%%%%%%%%%%%%%%%%%%%%%%%%%%%%%
\section{Summary and Conclusion}
This study addresses the fundamental friction between overparameterized TSFMs and the stringent predictive and latency constraints of quantitative financial environments. The core findings of this study are summarized.

\paragraph{Model Capacity Efficiency} A dense TSFM architecture ($\approx 2.68\,\text{M}$ parameters) achieves competitive forecasting accuracy against significantly larger foundation models, notably outperforming Moirai-13.8M. These results indicate that massive parameter scaling is not necessarily beneficial for low signal-to-noise ratio financial forecasting under fixed compute budgets. The observed $N^*$ is an empirical design choice under the tested optimizer settings, rather than an intrinsic capacity bound.
  
\paragraph{Inference Latency Reduction:} By integrating grouped-query attention into a lightweight decoder-only Transformer backbone, the proposed architecture substantially alleviates key-Value cache memory bandwidth bottlenecks during autoregressive decoding. Consequently, TailDART achieves up to a $48.1\%$ reduction in inference latency compared to standard multi-head attention architectures, reaching sub-millisecond execution speeds of $0.668\,\text{ms}$ per forecast step, thereby enabling seamless deployment within high-throughput, streaming financial infrastructure.
    
\paragraph{Empirical Multi-Asset Evaluation:} Extensive benchmarking across eight distinct asset classes across multiple temporal resolutions demonstrates that TailDART delivers consistently lower RMSE than Lag-Llama for all three: Daily, Weekly, and Monthly frequencies. Crucially, the model displays exceptional error resilience during extreme market shocks and high-volatility regimes, validating its structural robustness under heavy-tailed return distributions.

In summary, TailDART establishes that targeted architectural efficiency and capacity alignment offer a far more effective path for financial time-series foundation models than unconstrained parameter scaling. Future work will extend this framework toward real-time multi-asset portfolio execution and continuous online adaptation under rapid market regime shifts.

%%%%%%%%%%%%%%%%%%
\section*{Code Availability}
The source code, pre-trained model checkpoints, and configuration files to reproduce the experimental results of this work are publicly available on GitHub at \url{https://github.com/jjchat/taildart}.

%%%%%%%%%%%%%%%%%%
%acknowledgement
\section*{Acknowledgment}
This work was supported by the A*STAR's Industry Alignment Fund - Industry Collaboration Projects (IAF-ICP) Grant under Grant number H25-MCP3449.

%%%%%%%%%%%%%%%%%%%%%%%%%%%%%
\bibliographystyle{unsrt} %{ACM-Reference-Format}
\bibliography{references}

\end{document}